\documentclass[10pt, a4paper]{article}
\usepackage{lrec2022} 
\usepackage{multibib}
\newcites{languageresource}{Language Resources}
\usepackage{graphicx}
\usepackage{tabularx}
\usepackage{soul}

\usepackage{hyperref}       
\usepackage{url}            
\usepackage{booktabs}       %
\usepackage{amsfonts}       
\usepackage{nicefrac}
\usepackage{microtype} 
\usepackage{epstopdf}
\usepackage{amsmath}
\usepackage{hyperref}
\usepackage{xstring}
\usepackage{authblk}
\usepackage{amssymb}
\usepackage{graphicx}
\usepackage{multirow}
\usepackage{color}
\usepackage{subfigure}
\usepackage[capitalize,noabbrev]{cleveref}
\usepackage{pgfplots,pgfplotstable}
\pgfplotstableset{col sep=comma}
\pgfplotsset{compat=newest,}

\usepackage{titlesec}
\titleformat{\section}{\normalfont\large\bfseries\center}{\thesection.}{1em}{}
\titleformat{\subsection}{\normalfont\SmallTitleFont\bfseries\raggedright}{\thesubsection.}{1em}{}
\titleformat{\subsubsection}{\normalfont\normalsize\bfseries\raggedright}{\thesubsubsection.}{1em}{}
\renewcommand\thesection{\arabic{section}}
\renewcommand\thesubsection{\thesection.\arabic{subsection}}
\renewcommand\thesubsubsection{\thesubsection.\arabic{subsubsection}}

\usepackage{epstopdf}
\usepackage[utf8]{inputenc}

\usepackage{hyperref}
\usepackage{xstring}

\usepackage{color}

\usepackage{array}

\usepackage[normalem]{ulem} 

\usepackage{todonotes}

\title{Hausa Visual Genome: A Dataset for Multi-Modal English to Hausa Machine Translation}

\name{Idris Abdulmumin$^{1,6}$, Satya Ranjan Dash$^2$, Musa Abdullahi Dawud$^2$,\\ {\bf \large Shantipriya Parida$^3$, Shamsuddeen Hassan Muhammad$^{4,5}$, Ibrahim Sa'id Ahmad$^6$,} \\ {\bf \large Subhadarshi Panda$^7$, Ondřej Bojar$^8$, Bashir Shehu Galadanci$^6$, Bello Shehu Bello$^6$}}

\address{$^1$Department of Computer Science, Ahmadu Bello University, Zaria, Nigeria\\
	     $^2$School of Computer Applications, KIIT University, Bhubaneswar, India\\
	     $^3$Silo AI, Helsinki, Finland\\
	     $^4$ LIAAD - INESC TEC, $^5$Faculty of Sciences-University of Porto, Portugal\\
	     $^6$Faculty of Computer Science and Information Technology, Bayero University, Kano, Nigeria\\
	     $^7$Graduate Center, City University of New York, USA\\
	     $^8$Charles University, Faculty of Mathematics and Physics, \'{U}FAL, Prague, Czech Republic\\
         iabdulmumin@abu.edu.ng, sdashfca@kiit.ac.in,
         dawudmusa46@gmail.com,
         shantipriya.parida@silo.ai,\\
         \{shmuhammad.csc, isahmad.it, bsgaladanci.se, bsbello.cs\}@buk.edu.ng, spanda@gradcenter.cuny.edu,\\
         bojar@ufal.mff.cuni.cz
}

\abstract{
Multi-modal Machine Translation (MMT) enables the use of visual information to enhance the quality of translations. The visual information can serve as a valuable piece of context information to decrease the ambiguity of input sentences. Despite the increasing popularity of such a technique, good and sizeable datasets are scarce, limiting the full extent of their potential. Hausa, a Chadic language, is a member of the Afro-Asiatic language family. It is estimated that about 100 to 150 million people speak the language, with more than 80 million indigenous speakers. This is more than any of the other Chadic languages. Despite a large number of speakers, the Hausa language is considered low-resource in natural language processing (NLP). This is due to the absence of sufficient resources to implement most NLP tasks. While some datasets exist, they are either scarce, machine-generated, or in the religious domain. Therefore, there is a need to create training and evaluation data for implementing machine learning tasks and bridging the research gap in the language. This work presents the Hausa Visual Genome (HaVG), a dataset that contains the description of an image or a section within the image in Hausa and its equivalent in English. To prepare the dataset, we started by 
translating the English description of the images in the Hindi Visual Genome (HVG) into Hausa automatically. Afterward, the synthetic Hausa data was carefully post-edited considering the respective images. 
The dataset comprises 32,923 images and their descriptions that are divided into training, development, test, and challenge test set. The Hausa Visual Genome is the first dataset of its kind and can be used for Hausa-English machine translation, multi-modal research, and image description, among various other natural language processing and generation tasks.
 \\ \newline \Keywords{Natural Language Processing, Machine Translation, Hausa Language Resources, Low Resource Languages, Multi-modal dataset, Visual Genome} }

\begin{document}

\maketitleabstract


\section{Introduction}\label{intro}

Machine translation is the use of a computer to automatically generate the equivalent of a given source text in a language that is different from the original language. While Neural Machine Translation (NMT) \cite{Bahdanau2015,attention:vaswani:et:al,Gehring2017} has revolutionized automatic translation, the absence of sufficient training data in many languages has limited the benefits of such systems to a few languages rich in resources, although at least some treatment is possible even for low-resource languages
\cite{Sennrich2019}.

Multi-modal Machine Translation (MMT) enables the use of visual information to improve the translation quality, supplementing the missing context and providing cues to the machine translation system for better disambiguation. Despite the increasing popularity of multi-modal techniques, sufficiently large and clean datasets are scarce to fully benefit from the potential. For languages with such data, various approaches have been proposed, demonstrating their usability in improving translation quality, e.g., see \cite{Krishna2017,Lin2020,Long2021,Liu2021}.

\begin{figure*}[ht]
\begin{minipage}[t]{0.49\textwidth}
\resizebox{\linewidth}{!}{%
\includegraphics[width=0.5\textwidth]{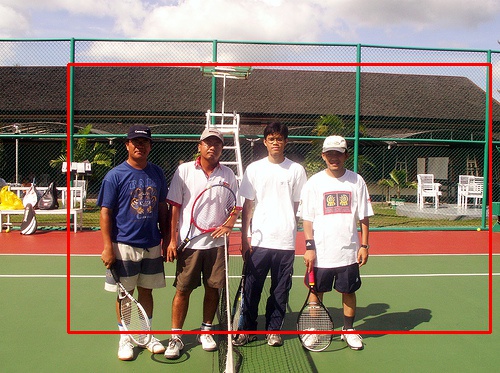}
}
~\textbf{English}: four men on court\\
~\textbf{Hausa}: \emph{maza hudu a \textbf{filin wasa}}\\
\hspace*{1em}\textbf{Gloss}: \emph{four men on a \textbf{playing field}}\\
~\textbf{MT}: \emph{maza hudu a \textbf{\textcolor{red}{kotu}}}\\
\hspace*{1em}\textbf{Gloss}: \emph{four men on a \textbf{\textcolor{red}{court}}}
\end{minipage}%
\hfill
\begin{minipage}[t]{0.03\textwidth}
\end{minipage}
\hfill
\begin{minipage}[t]{0.49\textwidth}
\resizebox{\linewidth}{!}{%
\includegraphics[width=0.5\textwidth]{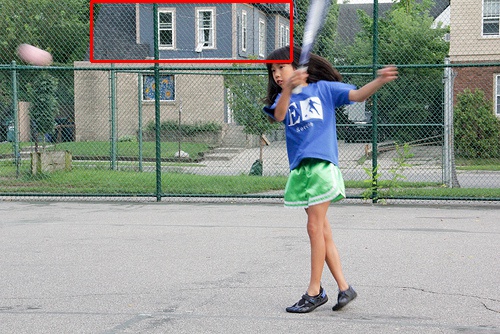}
}
~\textbf{English}: second story of house\\
~\textbf{Hausa}: \emph{\textbf{bene na biyu} na gida}\\
\hspace*{1em}\textbf{Gloss}: \emph{\textbf{second storey} of a house}\\
~\textbf{MT}: \emph{\textbf{\textcolor{red}{labarin}} gida \textbf{\textcolor{red}{na biyu}}}\\
\hspace*{1em}\textbf{Gloss}: \emph{\textbf{\textcolor{red}{story of second}} house}
\end{minipage}

\caption{Sample data from HaVG. The first translations (Hausa) are generated by Human Translators. The second translations (MT) are generated by a standard neural machine translation system, Google Translate. The wrong translations are in red font and bolded.
}
\label{fig:sample}

\end{figure*}

The images in \cref{fig:sample} present some examples where the absence of context allows to consider two different translations, where each is correct in a different setting. In the first one, the English word ``court'' is translated as \emph{kotu}, which is the Hausa word for a legal court. But the image illustrates that the men were standing on a [tennis] field. The absence of the word ``tennis'' misled the standard machine translation system and even many human translators into thinking that the former translation is required. The second 
example mentions a 
``story'' of a two-story house. The MT system translated the description as \emph{labarin}, meaning story (narrative), instead of the correct \emph{bene} (house story/storey). Without the picture, even human translators may 
make the same error given the very short and not quite correct English source.

Hausa is a Chadic language and a member of the Afro-Asiatic language family. Hausa is the most-spoken language in this family, with an estimate of about 100 to 150 million first-language and second-language speakers.\footnote{\url{https://www.herald.ng/full-list-hausa/}} The majority of these speakers are concentrated in the Northern part of Nigeria in cities such as Kano, Daura, Sokoto, Zaria, etc., and the Southern Niger Republic. The language is written in Arabic or Latin characters. The Arabic script is known as the \emph{Ajami} and was mostly used in the pre-colonial era, dating back to the 17th century \cite{Jaggar2006}. The language is nowadays written in the Latin script known as \emph{boko}.

Despite a large number of speakers and many written books, e.g., \newcite{Jaggar2006}, \newcite{Umar2013}, \newcite{Turner2021}, Hausa  is considered a low resource language in NLP. This is due to the absence of enough publicly available resources to implement most of the tasks in NLP. While some datasets exist, they are either scarce, machine-generated, or in the religious domain. This limits diversity, restricting the usage of trained models to very few domains. For tasks such as multi-modal translation, and image-to-text translation (image captioning), among others, there exist no training or evaluation data. For translation in the news domain, only an evaluation dataset exist\cite{FLORES2021}. Therefore, there is a need to create training and evaluation datasets for building machine learning models to help reduce the research gap between the low-resourced Hausa language and other languages.

This work, therefore, presents the Hausa Visual Genome (HaVG), a dataset that contains the description of an image or a section within the image in English and its equivalent in Hausa. The dataset was prepared by automatically translating the English description of the images in the Hindi Visual Genome (HVG) \cite{Parida2019}. The data is made of 32,923 images and their descriptions that are divided into training, development, test, and challenge test set. The machine-generated Hausa descriptions were then carefully post-edited taking into account the corresponding images. The HaVG is the first dataset of its kind in Hausa and can be used for Hausa-English machine translation, multi-modal research, and image description, among various other natural language processing and generation tasks.

The objective of the paper is two-fold:
\begin{enumerate}
     \item To describe the process of building the multimodal dataset for the Hausa language suitable for English-to-Hausa machine translation, image captioning, and multimodal research. 
    \item To demonstrate some sample use cases of the newly created multimodal dataset: HaVG.
\end{enumerate}

The rest of the paper is arranged as follows: \cref{available-hausa-data} presents the available datasets for NLP in the Hausa language. \cref{preparing-data} presents the processes of data collection and labeling. In \cref{sect_experiment}, we present some experiments and results on the application of the HaVG data. Finally, we conclude the work and provide directions for the future in \cref{sect_conclusion}.

\begin{figure*}[ht]
\begin{minipage}[t]{0.49\textwidth}
\resizebox{\linewidth}{!}{%
\includegraphics[width=0.5\textwidth]{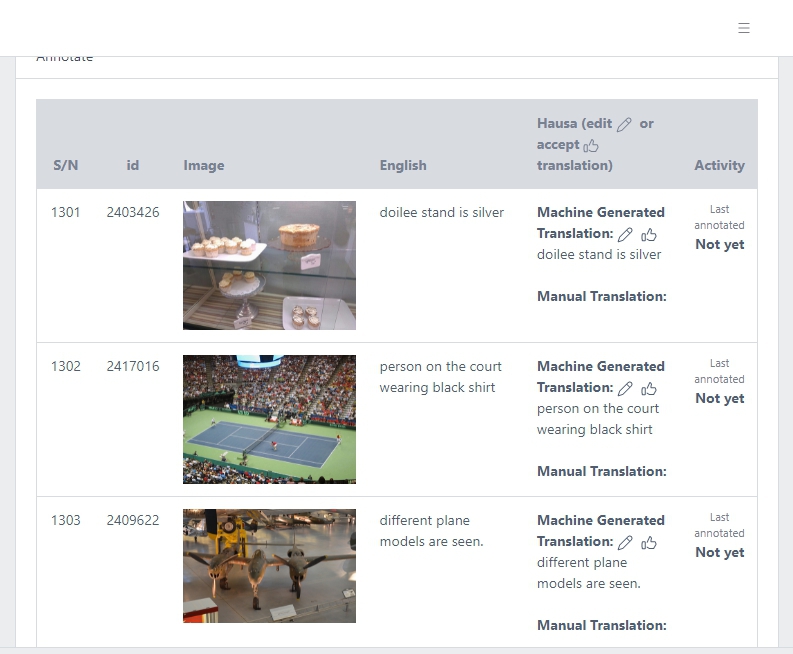}
}
\centering
\\~\textbf{a}: Preview\\
\end{minipage}%
\hfill
\begin{minipage}[t]{0.03\textwidth}
\end{minipage}
\hfill
\begin{minipage}[t]{0.49\textwidth}
\resizebox{\linewidth}{!}{%
\includegraphics[width=0.5\textwidth]{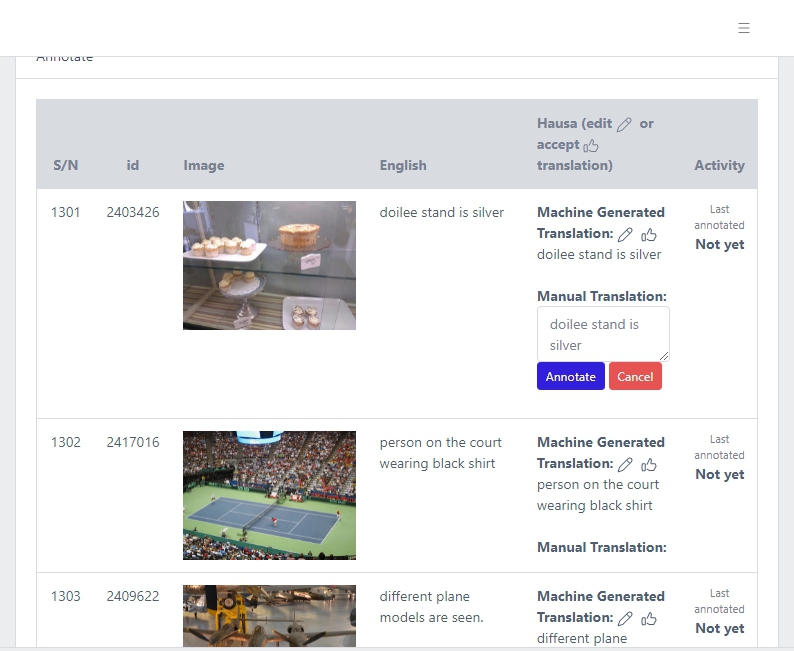}
}
\centering
\\~\textbf{b}: Annotate\\
\end{minipage}

\caption{Annotation web page showing images and their description. (a) To edit the machine translation, the pencil icon is clicked. To accept, the thumbs-up icon is clicked. (b) After clicking on the edit icon, a text area with the machine translation is displayed for post-editing.}
\label{fig:preview}

\end{figure*}

\section{Related Work}
\label{available-hausa-data}

While the Hausa language does not have any dataset for multimodal tasks, a few others have been created for other NLP tasks. \newcite{Abubakar2021} produced sentiment annotations of tweets and used them in their work. \newcite{Inuwa-Dutse2021} provided a pseudo-parallel corpus for machine translation. The Tanzil dataset\footnote{\url{https://opus.nlpl.eu/Tanzil.php}} \cite{TIEDEMANN12.463}, a translation of the Quran in many languages including Hausa, and the JW300\footnote{\url{https://opus.nlpl.eu/JW300.php}} \cite{jw300} are available for machine translation tasks. All of these data, though, are either not natural or strictly in the religious domain, limiting the accuracy or general applicability of the translation models trained on them. Apart from the FLoRes evaluation dataset \cite{FLORES2021} for machine translation tasks, which is not reflective of the domain of available training data, there exists no standard benchmark evaluation (test) sets that truly indicate the performance of natural language processing models to the best of our knowledge.

Resources for other NLP tasks in the language are also scarce. \newcite{Abdulmumin2019} provided two sets of word embeddings in Hausa for NLP. 
\newcite{Schlippe2012} and \newcite{Schultz2002} built a collection of transcribed speech resources for automatic speech recognition (ASR) and similar tasks in the language. \newcite{Tukur2019} trained a part-of-speech tagger for Hausa.

Initiatives such as Masakhane\footnote{\url{https://www.masakhane.io/}} and HausaNLP\footnote{\url{https://www.hausanlp.org/}} have started creating these data for Hausa and other African languages, most of which are considered low-resource and these will help in future NLP research and application in such languages.

\begin{table*}
\centering
\renewcommand{\arraystretch}{1.3}
\begin{tabular}{lrcrrrrr}
\hline

\multirow{2}{*}{Data} & \multirow{2}{*}{\#Sentences} & \multirow{2}{*}{Language} & \multicolumn{3}{c}{Word Stat.} & \multirow{2}{*}{\#Tokens} & \multirow{2}{*}{\#Vocab} \\
& & & max & min & avg & &  \\ \hline
\multirow{2}{*}{Training} & \multirow{2}{*}{28,930} & HA & 36 & 1 & 5.01 & 144,864 & 6,636 \\
& & EN & 29 & 1 & 5.09 & 147,219 & 7,046 \\ \hline
\multirow{2}{*}{Development Test} & \multirow{2}{*}{998} & HA & 14 & 1 & 4.99 & 4,978 & 1,167 \\
& & EN & 13 & 1 & 5.08 & 5,068 & 1,092 \\ \hline
\multirow{2}{*}{Evaluation Test} & \multirow{2}{*}{1,595} & HA & 17 & 1 & 4.99 & 7,952 & 1,478 \\
& & EN & 13 & 1 & 5.07 & 8,079 & 1,502 \\ \hline
\multirow{2}{*}{Challenge Test} & \multirow{2}{*}{1,400} & HA & 27 & 2 & 6.80 & 9,514 & 1,583 \\
& & EN & 18 & 2 & 6.01 & 8,411 & 1,461 \\ \hline

Total & 32,923 & -- & -- & -- & -- & -- & -- \\ \hline

\end{tabular}
\caption{Statistics of the Hausa Visual Genome dataset}
\label{tab:data-stat}
\end{table*}

\begin{table*}[ht]
    \centering
    \small
    \begin{tabular}{lccc}
        \hline
        Method & {D-Test BLEU} & {E-Test BLEU} & {C-Test BLEU} \\
        \hline
        Text-to-text translation & 31.3 & 46.7 & 17.7 \\
        Multimodal translation & 15.7 & 22.6 & 8.2 \\
        \hline
    \end{tabular}
    \caption{Results of text-only and multimodal translation on the HaVG dataset.}
    \label{tab:results}
\end{table*}

\section{Training and Evaluation Data}
\label{preparing-data}

\subsection{Data Collection}
The HaVG training and evaluation (development test and challenge test) data were produced by automatically translating the Hindi Visual Genome (HVG) and revising it as described below.

The HVG training, evaluation, and test dataset consist of randomly selected images and their descriptions from the Visual Genome (VG) corpus \cite{Krishna2017}. The HVG challenge test set was specifically sampled so that each sentence contains an English word that is lexically ambiguous when translated into Hindi. While the VG data contains multiple captions in English, with each caption representing a particular region in an image, the HVG data contains only a single random caption of a section in each image.

\subsection{Annotation}
To prepare the HaVG data, therefore, we implemented the following steps:
\begin{enumerate}
\item We use Google Translate\footnote{\url{https://translate.google.com/}} to translate all the available 32,923 HVG English captions into Hausa.

\item We developed a web-based annotation tool\footnote{\url{https://github.com/abumafrim/visual-genome-dataset-creation-tool}}
and hosted it locally to help with the post-editing of these translations. The web interface enables the annotator to edit the generated translations by showing them the image and the original caption side-by-side. See the illustration in \cref{fig:preview}.

\item We gave the machine translations of the captions to Hausa volunteers for post-editing. The translations of many of the unambiguous sentences were mostly found to be correct.

\item For a secondary check, we sampled 3,500 of the post-edited captions (representing about 10\% of the whole dataset) for manual verification. It was found that a small number of the sentences were found unedited even though there were obvious translation errors. The errors in these sentences were corrected by the verifiers.
\end{enumerate}

Some statistics in the annotated HaVG dataset are provided in \cref{tab:data-stat}. We used the NLTK punkt tokenizer \cite{bird2009natural} to estimate the statistics. The Hausa sentences of the HaVG were found to have 36 and 1 word in the longest and shortest sentences, respectively. The average sentence length ranges from 4.99 to 6.80 words per sentence, with the challenge test set statistically having longer sentences. The training set has a low type-token ratio (TTR) -- a measure of vocabulary variation or lexical richness of a text -- of 0.05. This is reflective of the restricted domain of the data as most of the sentences are in the sports domain, mainly tennis.

\section{Sample Applications of HaVG}
\label{sect_experiment}

\subsection{Text-Only Translation}
\label{subsect:text_only}
We used the Transformer model \cite{t2t_vaswani:et:al} as implemented in OpenNMT-py
\cite{opennmt}.\footnote{\url{http://opennmt.net/OpenNMT-py/quickstart.html}} Subword units were constructed using the word pieces algorithm \cite{google_word_pieces:melvin:etal}. Tokenization is handled automatically as part of the pre-processing pipeline of word pieces.

We generated a vocabulary of 32k subword types jointly for both the source and target languages, sharing it between the encoder and decoder. We used the Transformer base model \cite{t2t_vaswani:et:al}. We trained the model on a single GPU and followed the standard ``Noam'' learning rate decay,\footnote{\url{https://nvidia.github.io/OpenSeq2Seq/html/api-docs/optimizers.html}} see \newcite{attention:vaswani:et:al} or \newcite{transformer_tip:popel:et:al} for
 more details. Our starting learning rate was 0.2 and we used 8000 warm-up steps. The text-to-text translation results for the development (D-Test), dev (D-Test), test (E-Test), and challenge test (C-Test) are shown in \cref{tab:results}.
 
 In \cref{tab:text_multi_modal}, we present some examples where the text-only translation system was able to generate correct translations, although not the exact wording of the reference translations. The system  translated ``\textbf{stand}" as ``\emph{tsayuwa}" whereas the most appropriate translation should have been ``\emph{mazauni}" (with \emph{mazauni} meaning a place where something is kept while \emph{tsayuwa} means something/someone is in a standing position). The system also translated ``\textbf{block stone}" as ``\emph{dutse} (stone)", omitting ``block".
 
\begin{table*}[ht]
    \centering
    \small
    \begin{tabular}{crl}
        \hline
        \textbf{Image} & \textbf{Text}& \\ \hline\\
        \multirow{8}{*}{\includegraphics[height=3cm]{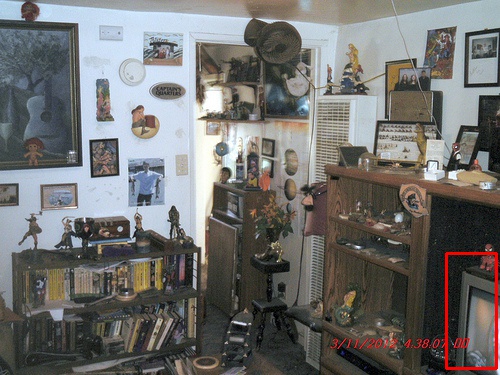} }&&\\
        & \textbf{Source} & Television in the tv stand. \\
        & \textbf{Reference} & Talabijin a cikin mazaunin talabijin \\ 
        & \textbf{\emph{Object Tags:}} & person, potted plant, book, tv, vase  \\
        & \textbf{Text-only} & Talabijin a cikin tsayuwa. \\
        & \textbf{\emph{Gloss.}} & Television in the \emph{standing}. \\
        & \textbf{Multi-modal} & Talabijin a cikin teburin tv \\
        & \textbf{\emph{Gloss.}} & Television in the tv table. \\ \hline\\
        \multirow{8}{*}{\includegraphics[height=3cm]{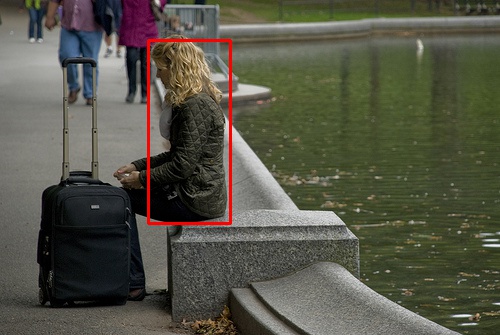} }&&\\
        & \textbf{Source} & woman sitting on a stone block \\
        & \textbf{Reference} & mace zaune a kan bulon dutse \\ 
        & \textbf{\emph{Object Tags:}} & person, suitcase, bench, remote \\
        & \textbf{Text-only} & mace zaune a kan dutse \\
        & \textbf{\emph{Gloss.}} & woman sitting on a stone \\
        & \textbf{Multi-modal} & mace zaune akan bangon dutse \\
        & \textbf{\emph{Gloss.}} & woman sitting on a stone wall \\ \hline
    \end{tabular}
    \caption{Text-only vs. Multi-modal Machine Translation.}
    \label{tab:text_multi_modal}
\end{table*}

\begin{table*}[ht]
    \centering
    \begin{tabular}{cccc}
    \hline
    \multirow{2}{*}{\textbf{Output}} & \multicolumn{3}{c}{\textbf{140 samples (10\%)}} \\
    \cline{2-4}
     & \textbf{Correct} & \textbf{Partially correct} & \textbf{Incorrect} \\
    \hline
    Text-only translation & 40 & 49 & 51 \\
    Multimodal translation & 13 & 39 & 88 \\
    \hline
    Multimodal translation resolves ambiguity & \multicolumn{3}{c}{14} \\   
    \hline
    Multimodal translation reasonable & \multicolumn{3}{c}{74} \\
    \hline
    \end{tabular}
    \caption{Comparison of outputs from various systems through manual evaluation of the challenge test set.}
    \label{tab:comparison_manual_evaluation}
\end{table*}

\subsection{Multimodal Translation}
\label{subsect:multimodal}

Multimodal translation involves utilizing the image modality in addition to the English text for translation to Hausa.
We take the multimodal neural machine translation approach using object tags derived from the image \cite{parida-etal-2021-multimodal}.
We first extract the list of (English) object tags for a given image using the pre-trained Faster R-CNN \cite{Ren_Faster_R_CNN} with ResNet101-C4 \cite{he2016deep} backbone.
We pick the top 10 object tags based on their confidence scores.
In cases where less than 10 object tags are detected, we consider all tags.

Next, the object tags are concatenated to the English sentence which needs to be translated to Hausa.
The concatenation is done using the special token `\#\#’ as the separator.
The separator is followed by comma-separated object tags.
Adding object labels enables the otherwise text-based model to utilize visual concepts which may not be readily available in the original sentence.
The English sentences along with the object tags are fed to the encoder of a text-to-text Transformer model.
The decoder generates the Hausa translations auto-regressively. We generated a vocabulary of 50k subwords 
for both source and target languages. Then we trained the Transformer base model using the ``Noam" learning rate decay. We used an initial learning rate of 2, dropout of 0.1, and 8000 warm-up steps.       
The results of the multimodal translation are shown in \cref{tab:results}.

The automatic evaluation indicates that the text-only translation performs better on both the evaluation and challenge test sets when compared to the multimodal translation. However, upon manual inspection of the outputs, we observed instances where the multimodal system was able to resolve ambiguity and generate a more appropriate translation of the given source sentence, see \cref{tab:text_multi_modal} for some examples. The performance is strikingly lower on the challenge test set compared to the evaluation set in both setups. We performed a manual evaluation on a sample of this data to investigate the reason for this low performance.

About 10\% of the translations of the challenge test set by both the text-only and multimodal systems were sampled and manually evaluated to assess the quality of the generated sentences. We categorized these sentences as either \textbf{correct}, \textbf{partially correct}, or \textbf{incorrect}. We also checked instances where the multimodal system is not only correct (or partially correct) but was also able to resolve ambiguity. Lastly, we checked whether the sentences generated by the multimodal system are reasonable or not, i.e. whether they generally capture the original meaning. The results of this evaluation are provided in \cref{tab:comparison_manual_evaluation}.

While the multimodal system was found to be half as accurate compared to the text-only model, it was able to resolve ambiguity in about 10\% of the sampled data.
Finally, we observe that the annotation for ``reasonable'' translations (i.e. whether the meaning is ``generally captured'' is apparently much more permissive that the annotation for correctness: a substantial amount of the generated text (74 items, i.e. 53\%) was found to be reasonable even though only about 37\% of the sentences are either correct or partially correct translations of the source sentences. This detailed analysis nevertheless confirmed that the multi-modal system produces overall worse translations, perhaps confused by the automatic object captions.

\begin{figure*}[ht]
\centering
\includegraphics[width=0.9\linewidth]{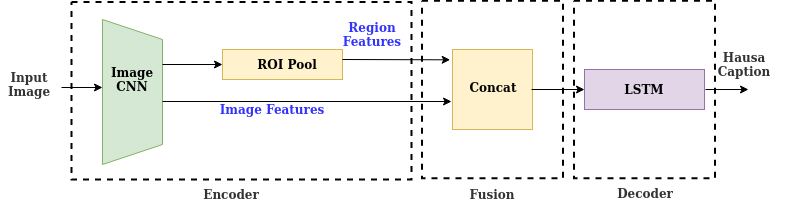}

\caption{Architecture of the region-specific image caption generator.}
\label{fig:arch_caption}
\end{figure*}

\subsection{Image Caption Generation}
\label{subsect:caption}

To generate the Hausa captions, we followed  \newcite{parida2021nlphut} who proposed a region-specific image captioning method through the fusion of the encoded features of the region and the complete image. The model consists of three modules -- an encoder, fusion, and decoder -- as shown in \cref{fig:arch_caption}.

\begin{table*}[t]
    \centering
    \small
    \begin{tabular}{lccc}
        \hline
        Method & {D-Test BLEU} & {E-Test BLEU} & {C-Test BLEU} \\
        \hline
        Image captioning & 2.6 & 3.1 & 0.7 \\
        \hline
    \end{tabular}
    \caption{Results of image caption generation on the HaVG dataset.}
    \label{tab:results_caption}
\end{table*}

\paragraph{Image encoder}
In the proposed approach, the features of the entire image, as well as features of the sub-region, are considered to train the model. The features from the corresponding regions are extracted through Region of Interest (RoI) pooling \cite{fast_rcnn2015}. Specifically, the feature vector is the output of the fourth block of ResNet-50 in our experiments.
It is a 2048-dimensional vector for both the image and the sub-region.
We keep the image encoder module non-trainable.
In other words, it is used as a feature extractor.

\paragraph{Fusion module}
While the region-level features capture details of the region (objects) to be described, the image-level features provide an overall context. To generate a meaningful caption, both need to be fused appropriately.
We obtain the final feature vector by simple concatenation of features from the region and features from the entire image. The concatenation resulted in a 4096-dimensional vector.

\paragraph{LSTM decoder}
The concatenated feature vector is passed through a linear layer to project it into a 128-dimensional vector which is then fed as input to an LSTM decoder as the first time step. The decoder generates the tokens of the caption autoregressively using a greedy search approach.
A single-layer LSTM is used and its hidden size is set to 256.
The dropout is set to 0.3.
While the image encoder module is non-trainable, the LSTM decoder module is trainable.
During training, the cross-entropy loss is minimized, which is computed using the output logits and the tokens in the gold caption.
Weights are optimized using the Adam optimizer \cite{Adam} with an initial learning rate of 0.0001.
Training is halted when the validation loss does not improve for 10 consecutive epochs.

The results of the image captioning in terms of BLEU scores are shown in \cref{tab:results_caption}.
We observe that the BLEU scores of the generated image captions are much lower than the translation-based captions.

This is not very surprising because automatic captioning is free to choose a very different aspect of the image or use wording very different from the reference caption. BLEU only checks for n-gram overlap between the caption and the reference.
Therefore, we perform a manual evaluation to further analyze the performance of the image caption generation model.

\pgfplotstableread[row sep=\\,col sep=&]{
    label       & count \\
    Match OOI   & 42  \\
    Match ROI   & 10  \\
    Other Region     & 44 \\
    Wrong       & 45 \\
}\captionManualEvalData

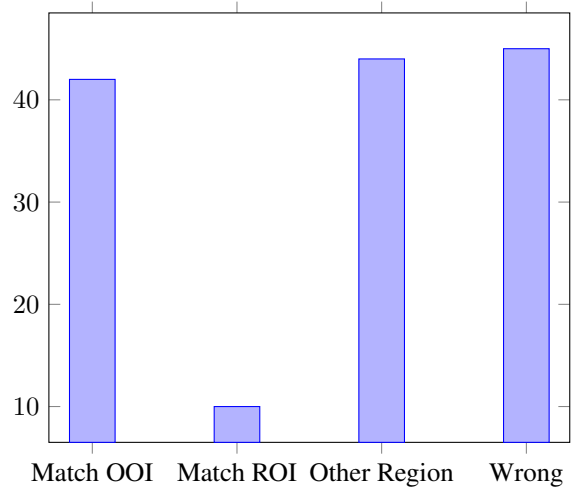
\begin{figure}[t]
    \begin{tikzpicture}
        \begin{axis}[
                ybar,
                symbolic x coords={Match OOI,Match ROI,Other Region,Wrong},
                xtick=data,
                bar width = .6cm,
            ]
            \addplot table[x=label,y=count]{\captionManualEvalData};
        \end{axis}
    \end{tikzpicture}
\caption{Manual Evaluation of Sampled Generated Captions.}
\label{fig:caption_manual_eval}
\end{figure}

\begin{table*}[!ht]
    \small
    \begin{tabular}{p{1.1cm}p{6cm}|p{1.1cm}p{6cm}}
        \hline \hline
        \multicolumn{4}{|c|}{\textbf{Match OOI}} \\ \hline \hline
        \multicolumn{2}{c|}{\includegraphics[height=3.4cm]{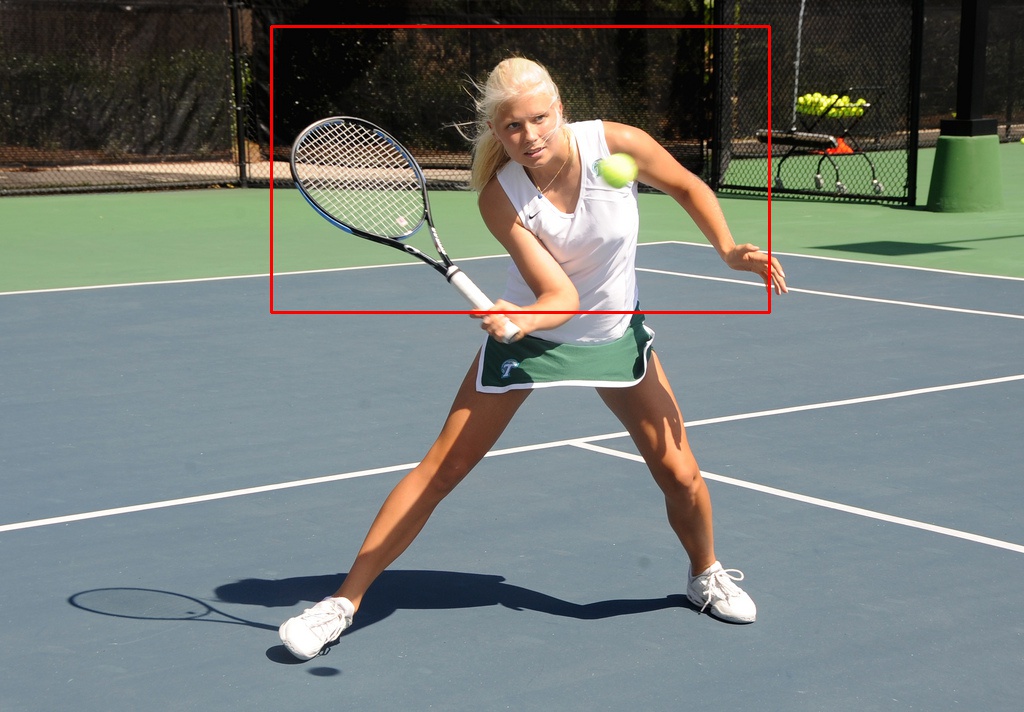}} & \multicolumn{2}{c}{\includegraphics[height=3.4cm]{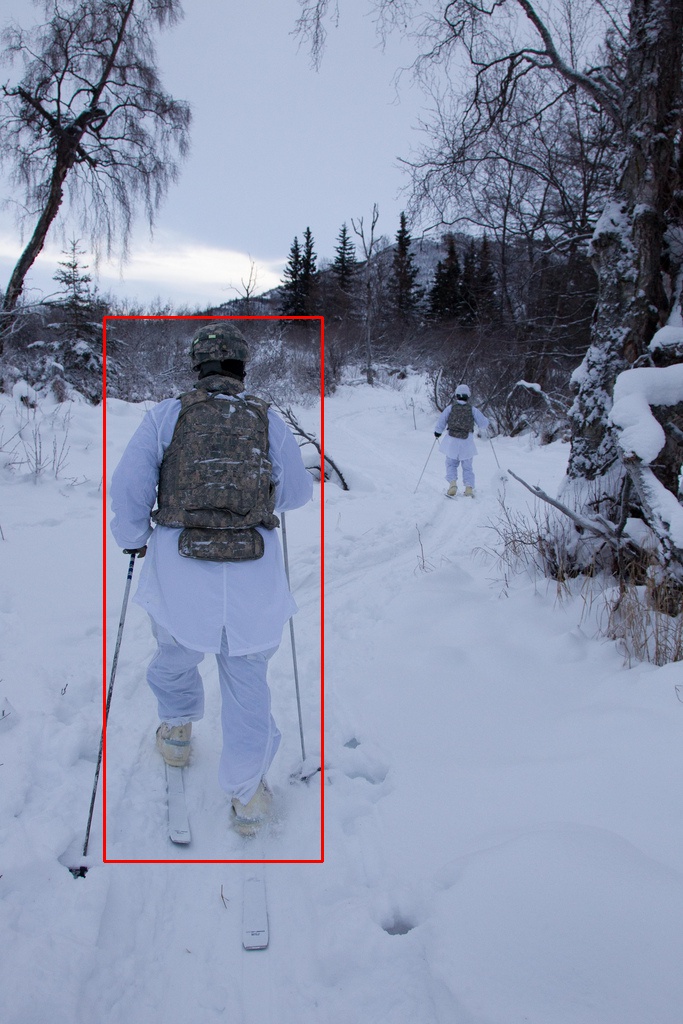}} \\
        \hline
        \textbf{Reference} & Wata yarinya a filin wasan tanis tana shirin buga kwallon & \textbf{Reference} & mutum na biyu yana gudun kan dusar kankara \\
        \textbf{Gloss} & A girl on the tennis court is preparing to hit the ball & \textbf{Gloss} & second man skiing in snow \\ \hline
        \textbf{Model} & mutumin da ke wasan tennis & \textbf{Model} & mutum yana kan kankara \\
        \textbf{Gloss} & the person playing tennis & \textbf{Gloss} & person is on snow 
        \\\hline \hline
        \multicolumn{2}{|c}{\textbf{Match ROI}} & \multicolumn{2}{|c|}{\textbf{Other Region}} \\ \hline \hline
        \multicolumn{2}{c|}{\includegraphics[height=3.4cm]{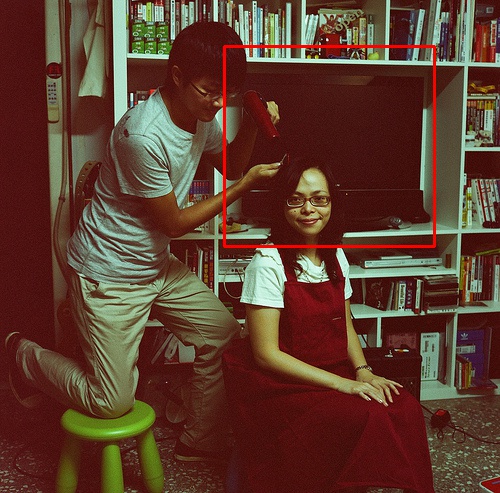}} & \multicolumn{2}{c}{\includegraphics[height=3.4cm]{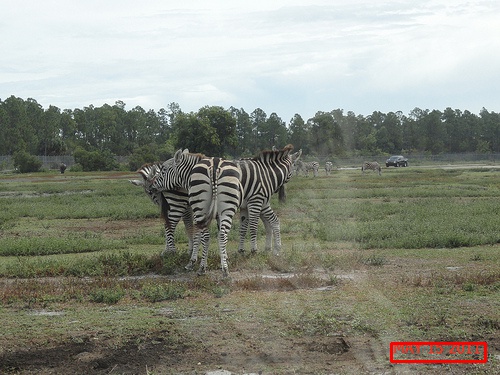}} \\
        \hline
        \textbf{Reference} & TALABIJIN a tsaye. & \textbf{Reference} & hatimin kwanan wata a kusurwar hoton \\
        \textbf{Gloss} & TV on the stand. & \textbf{Gloss} & the date stamp in the corner of the pic \\ \hline
        \textbf{Model} & mutum yana sanye da tabarau & \textbf{Model} & alfadari a cikin ciyawa \\
        \textbf{Gloss} & person wearing glasses & \textbf{Gloss} & zebra in the grass 
        \\\hline \hline
        \multicolumn{4}{|c|}{\textbf{Wrong}} \\ \hline \hline
        \multicolumn{2}{c|}{\includegraphics[height=3.4cm]{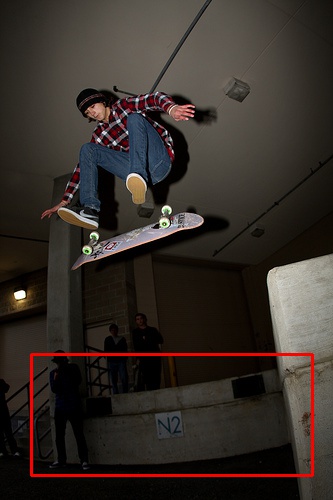}} & \multicolumn{2}{c}{\includegraphics[height=3.4cm]{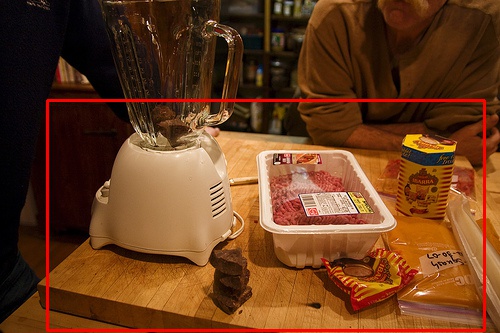}} \\
        \hline
        \textbf{Reference} & babban siminti & \textbf{Reference} & wani bulon katako da ke zaune a kan tebur \\
        \textbf{Gloss} & large cement block & \textbf{Gloss} & a wooden block sitting on the table \\ \hline
        \textbf{Model} & mutum yana kan kankara & \textbf{Model} & wani mutum yana cin abinci \\
        \textbf{Gloss} & person is on snow & \textbf{Gloss} & a person eating food 
        \\\hline \hline
      
    \end{tabular}
    \caption{Manual classification of the qualities of sampled region of interest captions taken from the challenge dataset.}
    \label{tab:caption_manual_eval_example}
\end{table*}

\subsubsection{Manual Evaluation}
\label{sec:manual:evaluation}
A sample of about 10\% of the generated captions was manually evaluated and categorized into the following classes:
\paragraph{Match OOI} for captions that describe the object of interest provided in the reference caption, exactly or closely.
\paragraph{Match ROI} for captions that describe a different object within the region of interest.
\paragraph{Other Region} for captions that describe an object in the image that is outside the region of interest.
\paragraph{Wrong} for captions that do not describe any object in the associated image.

\cref{fig:caption_manual_eval} presents the result of the manual evaluation of the sampled machine-generated captions. From the evaluated sample, it was observed that about 68\% of the generated captions correctly describe an object in the image. Of this number, about 54\% of the captions describe an object in the region of interest. However, most of the descriptions, although correct, do not match the description given in the reference caption (our evaluation does not quantify this aspect.)

This explains the low BLEU scores reported in \cref{tab:results}. A more appropriate metric may be needed, therefore, to correctly measure the performance of such systems.

In \cref{tab:caption_manual_eval_example}, we provide examples of each of these manual evaluation classes.

\section{Conclusion and Future Work}
\label{sect_conclusion}

We present the HaVG, the multimodal dataset suitable for English$\rightarrow$Hausa machine translation, image captioning, and multimodal research. 

The dataset is freely available for research and non-commercial usage under a Creative Commons Attribution-NonCommercial-ShareAlike 4.0 License\footnote{\url{https://creativecommons.org/licenses/by-nc-sa/4.0/}} at: \url{http://hdl.handle.net/11234/1-4749}.

In future versions of the HaVG, we plan to create the dataset from scratch without relying on an initial MT system and post-editing.
Other future works include \textit{i)} organizing a shared task using the HaVG, \textit{ii)} extending the HaVG corpus for Visual Question Answering (VQA).

\section{Acknowledgements}
This work has received funding from the grant 19-26934X (NEUREM3) of the Czech Science Foundation, and has also been supported by the Ministry of Education, Youth and Sports of the Czech Republic, Project No. LM2018101 LINDAT/CLARIAH-CZ. This work is also financed by National Funds through the Portuguese funding agency, FCT - Fundação para a Ciência e a Tecnologia, within project LA/P/0063/2020.

\section{Bibliographical References}\label{reference}

\bibliographystyle{lrec2022-bib}
\bibliography{lrec2022-example}

\begin{thebibliography}{}

\bibitem[\protect\citename{Abdulmumin and Galadanci}2019]{Abdulmumin2019}
Abdulmumin, I. and Galadanci, B.~S.
\newblock (2019).
\newblock {hauWE: Hausa Words Embedding for Natural Language Processing}.
\newblock In {\em 2019 2nd International Conference of the IEEE Nigeria
  Computer Chapter (NigeriaComputConf)}, pages 1--6. IEEE.

\bibitem[\protect\citename{Abubakar \bgroup et al.\egroup }2021]{Abubakar2021}
Abubakar, A.~I., Roko, A., Bui, A.~M., and Saidu, I.
\newblock (2021).
\newblock {An Enhanced Feature Acquisition for Sentiment Analysis of English
  and Hausa Tweets}.
\newblock {\em International Journal of Advanced Computer Science and
  Applications (IJACSA)}, 12(9):102--110.

\bibitem[\protect\citename{Agi{\'c} and Vuli{\'c}}2019]{jw300}
Agi{\'c}, {\v{Z}}. and Vuli{\'c}, I.
\newblock (2019).
\newblock {JW}300: A wide-coverage parallel corpus for low-resource languages.
\newblock In {\em Proceedings of the 57th Annual Meeting of the Association for
  Computational Linguistics}, pages 3204--3210, Florence, Italy, July.
  Association for Computational Linguistics.

\bibitem[\protect\citename{Bahdanau \bgroup et al.\egroup }2015]{Bahdanau2015}
Bahdanau, D., Cho, K., and Bengio, Y.
\newblock (2015).
\newblock {Neural Machine Translation by Jointly Learning to Align and
  Translate}.
\newblock In Yoshua Bengio et~al., editors, {\em 3rd International Conference
  on Learning Representations, {\{}ICLR{\}} 2015, San Diego, CA, USA, May 7-9,
  2015, Conference Track Proceedings}.

\bibitem[\protect\citename{Bird \bgroup et al.\egroup }2009]{bird2009natural}
Bird, S., Klein, E., and Loper, E.
\newblock (2009).
\newblock {\em Natural language processing with Python: analyzing text with the
  natural language toolkit}.
\newblock " O'Reilly Media, Inc.".

\bibitem[\protect\citename{Gehring \bgroup et al.\egroup }2017]{Gehring2017}
Gehring, J., Michael, A., Grangier, D., Yarats, D., and Dauphin, Y.~N.
\newblock (2017).
\newblock {Convolutional Sequence to Sequence Learning}.
\newblock In Doina Precup et~al., editors, {\em Proceedings of the 34th
  International Conference on Machine Learning}, volume~70, pages 1243--1252,
  Sydney, Australia. PMLR.

\bibitem[\protect\citename{Girshick}2015]{fast_rcnn2015}
Girshick, R.
\newblock (2015).
\newblock Fast r-cnn.
\newblock In {\em Proceedings of the IEEE International Conference on Computer
  Vision (ICCV)}, pages 1440--1448.

\bibitem[\protect\citename{Goyal \bgroup et al.\egroup }2021]{FLORES2021}
Goyal, N., Gao, C., Chaudhary, V., Chen, P.-J., Wenzek, G., Ju, D., Krishnan,
  S., Ranzato, M., Guzman, F., and Fan, A.
\newblock (2021).
\newblock The flores-101 evaluation benchmark for low-resource and multilingual
  machine translation.

\bibitem[\protect\citename{He \bgroup et al.\egroup }2016]{he2016deep}
He, K., Zhang, X., Ren, S., and Sun, J.
\newblock (2016).
\newblock Deep residual learning for image recognition.
\newblock In {\em Proceedings of the IEEE conference on computer vision and
  pattern recognition}, pages 770--778.

\bibitem[\protect\citename{Inuwa-Dutse}2021]{Inuwa-Dutse2021}
Inuwa-Dutse, I.
\newblock (2021).
\newblock {The first large scale collection of diverse Hausa language
  datasets}.
\newblock {\em arXiv e-prints}.

\bibitem[\protect\citename{Jaggar}2006]{Jaggar2006}
Jaggar, P.~J.
\newblock (2006).
\newblock {Hausa}.
\newblock {\em Elsevier Ltd}, pages 222--225.

\bibitem[\protect\citename{Johnson \bgroup et al.\egroup
  }2017]{google_word_pieces:melvin:etal}
Johnson, M., Schuster, M., Le, Q.~V., Krikun, M., Wu, Y., Chen, Z., Thorat, N.,
  Vi{\'e}gas, F., Wattenberg, M., Corrado, G., Hughes, M., and Dean, J.
\newblock (2017).
\newblock Google's multilingual neural machine translation system: Enabling
  zero-shot translation.
\newblock {\em Transactions of the Association for Computational Linguistics},
  5:339--351.

\bibitem[\protect\citename{Kingma and Ba}2014]{Adam}
Kingma, D.~P. and Ba, J.
\newblock (2014).
\newblock Adam: A method for stochastic optimization.
\newblock cite arxiv:1412.6980Comment: Published as a conference paper at the
  3rd International Conference for Learning Representations, San Diego, 2015.

\bibitem[\protect\citename{Klein \bgroup et al.\egroup }2017]{opennmt}
Klein, G., Kim, Y., Deng, Y., Senellart, J., and Rush, A.~M.
\newblock (2017).
\newblock Open{NMT}: Open-source toolkit for neural machine translation.
\newblock In {\em Proc. ACL}.

\bibitem[\protect\citename{Krishna \bgroup et al.\egroup }2017]{Krishna2017}
Krishna, R., Zhu, Y., Groth, O., Johnson, J., Hata, K., Kravitz, J., Chen, S.,
  Kalantidis, Y., Li, L.-J., Shamma, D.~A., Bernstein, M.~S., and Fei-Fei, L.
\newblock (2017).
\newblock {Visual Genome: Connecting Language and Vision Using Crowdsourced
  Dense Image Annotations}.
\newblock {\em International Journal of Computer Vision}, 123(1):32--73.

\bibitem[\protect\citename{Lin \bgroup et al.\egroup }2020]{Lin2020}
Lin, H., Meng, F., Su, J., Yin, Y., Yang, Z., Ge, Y., Zhou, J., and Luo, J.
\newblock (2020).
\newblock {Dynamic Context-guided Capsule Network for Multimodal Machine
  Translation}.
\newblock In {\em Proceedings of the 28th ACM International Conference on
  Multimedia}, pages 1320--1329, New York, NY, USA. Association for
  Computational Linguistics.

\bibitem[\protect\citename{Liu \bgroup et al.\egroup }2021]{Liu2021}
Liu, P., Cao, H., and Zhao, T.
\newblock (2021).
\newblock {Gumbel-Attention for Multi-modal Machine Translation}.
\newblock {\em CoRR}.

\bibitem[\protect\citename{Long \bgroup et al.\egroup }2021]{Long2021}
Long, Q., Wang, M., and Li, L.
\newblock (2021).
\newblock {Generative Imagination Elevates Machine Translation}.
\newblock In {\em 2021 Conference of the North American Chapter of the
  Association for Computational Linguistics: Human Language Technologies},
  pages 5738--5748. Association for Computational Linguistics.

\bibitem[\protect\citename{Parida \bgroup et al.\egroup }2019]{Parida2019}
Parida, S., Bojar, O., and Dash, S.~R.
\newblock (2019).
\newblock {Hindi Visual Genome: A Dataset for Multi-Modal English to Hindi
  Machine Translation}.
\newblock {\em Computaci{\'{o}}n y Sistemas}, 23(4).

\bibitem[\protect\citename{Parida \bgroup et al.\egroup
  }2021a]{parida-etal-2021-multimodal}
Parida, S., Panda, S., Biswal, S.~P., Kotwal, K., Sen, A., Dash, S.~R., and
  Motlicek, P.
\newblock (2021a).
\newblock Multimodal neural machine translation system for {E}nglish to
  {B}engali.
\newblock In {\em Proceedings of the First Workshop on Multimodal Machine
  Translation for Low Resource Languages (MMTLRL 2021)}, pages 31--39, Online
  (Virtual Mode), September. INCOMA Ltd.

\bibitem[\protect\citename{Parida \bgroup et al.\egroup
  }2021b]{parida2021nlphut}
Parida, S., Panda, S., Kotwal, K., Dash, A.~R., Dash, S.~R., Sharma, Y.,
  Motlicek, P., and Bojar, O.
\newblock (2021b).
\newblock Nlphut’s participation at wat2021.
\newblock In {\em Proceedings of the 8th Workshop on Asian Translation
  (WAT2021)}, pages 146--154.

\bibitem[\protect\citename{Popel and Bojar}2018]{transformer_tip:popel:et:al}
Popel, M. and Bojar, O.
\newblock (2018).
\newblock Training tips for the transformer model.
\newblock {\em The Prague Bulletin of Mathematical Linguistics}, 110(1):43--70.

\bibitem[\protect\citename{Ren \bgroup et al.\egroup }2015]{Ren_Faster_R_CNN}
Ren, S., He, K., Girshick, R., and Sun, J.
\newblock (2015).
\newblock Faster r-cnn: Towards real-time object detection with region proposal
  networks.
\newblock In {\em Proceedings of the 28th International Conference on Neural
  Information Processing Systems - Volume 1}, NIPS'15, page 91–99, Cambridge,
  MA, USA. MIT Press.

\bibitem[\protect\citename{Schlippe \bgroup et al.\egroup }2012]{Schlippe2012}
Schlippe, T., Djomgang, E. G.~K., Vu, N.~T., Ochs, S., and Schultz, T.
\newblock (2012).
\newblock {Hausa Large Vocabulary Continuous Speech Recognition}.
\newblock In {\em Spoken Language Technologies for Under-Resourced Languages}.

\bibitem[\protect\citename{Schultz}2002]{Schultz2002}
Schultz, T.
\newblock (2002).
\newblock {GlobalPhone: A Multilingual Speech and Text Database Developed at
  Karlsruhe University}.
\newblock In {\em Seventh International Conference on Spoken Language
  Processing}, pages 345---348.

\bibitem[\protect\citename{Sennrich and Zhang}2019]{Sennrich2019}
Sennrich, R. and Zhang, B.
\newblock (2019).
\newblock {Revisiting Low-Resource Neural Machine Translation: A Case Study}.
\newblock In {\em Proceedings of the 57th Annual Meeting of the Association for
  Computational Linguistics}, pages 211--221, Stroudsburg, PA, USA. Association
  for Computational Linguistics.

\bibitem[\protect\citename{Tiedemann}2012]{TIEDEMANN12.463}
Tiedemann, J.
\newblock (2012).
\newblock Parallel data, tools and interfaces in opus.
\newblock In Nicoletta Calzolari~(Conference Chair), et~al., editors, {\em
  Proceedings of the Eight International Conference on Language Resources and
  Evaluation (LREC'12)}, Istanbul, Turkey, may. European Language Resources
  Association (ELRA).

\bibitem[\protect\citename{Tukur \bgroup et al.\egroup }2019]{Tukur2019}
Tukur, A., Umar, K., and Muhammad, A.~S.
\newblock (2019).
\newblock {Tagging Part of Speech in Hausa Sentences}.
\newblock In {\em 2019 15th International Conference on Electronics, Computer
  and Computation (ICECCO)}, pages 1--6.

\bibitem[\protect\citename{Turner}2021]{Turner2021}
Turner, T.~D.
\newblock (2021).
\newblock {Hausa Songs in Algeria: sounds of trans-Saharan continuity and
  rupture}.
\newblock {\em The Journal of North African Studies}, pages 1--29, may.

\bibitem[\protect\citename{Umar}2013]{Umar2013}
Umar, M.~S.
\newblock (2013).
\newblock {Hausa Traditional Political Culture, Islam, and Democracy:
  Historical Perspectives on Three Political Traditions}.
\newblock In {\em Democracy and Prebendalism in Nigeria}, pages 177--200.
  Palgrave Macmillan US, New York.

\bibitem[\protect\citename{Vaswani \bgroup et al.\egroup
  }2017]{attention:vaswani:et:al}
Vaswani, A., Shazeer, N., Parmar, N., Uszkoreit, J., Jones, L., Gomez, A.~N.,
  Kaiser, {\L}., and Polosukhin, I.
\newblock (2017).
\newblock Attention is all you need.
\newblock In {\em Advances in Neural Information Processing Systems}, pages
  5998--6008.

\bibitem[\protect\citename{Vaswani \bgroup et al.\egroup
  }2018]{t2t_vaswani:et:al}
Vaswani, A., Bengio, S., Brevdo, E., Chollet, F., Gomez, A., Gouws, S., Jones,
  L., Kaiser, {\L}., Kalchbrenner, N., Parmar, N., Sepassi, R., Shazeer, N.,
  and Uszkoreit, J.
\newblock (2018).
\newblock Tensor2tensor for neural machine translation.
\newblock In {\em Proc. of AMTA (Volume 1: Research Papers)}, pages 193--199.

\end{thebibliography}

\end{document}